# Multibiometric: Feature Level Fusion Using FKP Multi-Instance biometric

Harbi AlMahafzah[1], Mohammad Imran[2], and H.S. Sheshadri[3]

[1] P.E.T. Foundation Research, University of Mysore,
Mandya, India
hmahafzah@hotmail.com
[2] Department of Study in Computer Science, University of Mysore,
Mysore, India
emraangi@gmail.com
[3] Department E & C Engg, P.E.S College of Engineering, V.T.U,
Mandya, India
hssheshadri@hotmail.com

**Abstract.** This paper proposed the use of multi-instance feature level fusion as a means to improve the performance of Finger Knuckle Print (FKP) verification. A log-Gabor filter has been used to extract the image local orientation information, and represent the FKP features. Experiments are performed using the FKP database, which consists of 7,920 images. Results indicate that the multi-instance verification approach outperforms higher performance than using any single instance. The influence on biometric performance using feature level fusion under different fusion rules have been demonstrated in this paper.

## 1 Introduction

The need for reliable user authentication techniques has increased in the wake of heightened concerns about security and rapid advancements in networking, communication, and mobility. A wide variety of applications require reliable verification schemes to confirm the identity of an individual requesting their service. Traditional authentication methods using passwords (knowledge-based security) and ID cards (token based security) are commonly used to restrict access to a variety of systems. However these systems are vulnerable to attacked and security can be easily breached. The emergence of biometrics technologies is replacing the traditional methods as it has addressed the problems that plague these systems.

Biometric refers to the automatic recognition of individuals based on their physiological and behavioral characteristics. Biometrics systems are commonly classified into two categories: physiological biometrics and behavioral biometrics. Physiological biometrics (fingerprint, iris, retina, hand geometry, face, etc) use measurements from the human body. Behavioral biometrics (signature, keystrokes, voice, etc) use dynamics measurements based on human actions [1][3]. These systems are based on pattern recognition methodology, which follows the acquisition of the biometric data by building a biometric feature set, and comparing versus a pre-stored template pattern. These are unimodal which rely on the evidence of a single source of information for authentication, which have to contend with a variety of problems such as (noise in sensed data, intra-class variations, and inter-class similarities, etc). It is now apparent that a single biometric is not sufficient to meet the variety of requirements imposed by several large scale authentication systems. Possible solutions to compensate for the false classification problem due to intra-class variability and inter-class similarity can be found in the fusion of biometric systems or experts [8] which refers as Multibiometric.

The Multibiometric systems can offer substantial improvement in the matching accuracy of a biometric system depending upon the information being combined and the fusion methodology adopted [1] as follows; **Multi sensor**: Multiple sensors can be used to collect the same biometric. **Multi-modal**: Multiple biometric modalities can be collected from the same individual, e.g. fingerprint and face, which requires different sensors. **Multi-sample**: Multiple readings of the same biometric are collected during the enrolment and/or recognition phases, e.g. a number of fingerprint readings are taken from the same finger. **Multiple algorithms**: Multiple algorithms for feature extraction and matching are used on the same biometric sample. **Multi-instance**: -which this paper concentrate on-means the use of the same type of raw biometric sample and processing on multiple instances of similar body parts, (such as two fingers, or two irises) also been referred to as multi-unit systems in the literature [2].

Multi-instance systems can be cost-effective since a single sensor is used to acquire the multi-unit data in a sequential fashion, and these systems generally do not necessitate the introduction of new sensors nor do they entail the development of new feature extraction and matching algorithms. Multi-instance systems are especially beneficial to users whose biometric traits cannot be reliably captured due to inherent problems. Multi-instance systems are often necessary in applications where the size of the system database (i.e., the number of enrolled individuals) is very large [2] (FBI's database currently has 50 million ten-print images, and multiple fingers provide additional discriminatory information). Combination of multi-instances can improve the performance of the biometric system.

## 2   Related Works

Lin Zhang et al.[4] proposed an effective FKP recognition scheme by extracting and assembling local and global features of FKP images. The experimental results conducted on FKP database demonstrate that the proposed local–global information combination scheme could significantly improve the recognition accuracy obtained by either local or global information are lead to promising performance of an FKP-based personal authentication system. The authors experimental results conducted on FKP database indicate that the proposed scheme could achieve much better performance in terms of EER and the decidability index than the other state-of- the-art competitors.

T.C. Faltemier et al. [6] proposed a multi-instance enrollment for face recognition as a means to improve the performance of 3D face recognition. The authors show that using multiple images to enroll a person in a gallery can improve the overall performance of a biometric system. The authors demonstrated that when using multiple images to enroll a person, sampling from different expressions improves performance over sampling only the same expression. Tobias Scheidat, et al. [8] proposed a fusion of two instances of the same semantic, where semantics are alternative handwritten contents such as numbers or sentences, in addition to commonly used signature. In order to fuse two instances of one semantic, a biometric authentication is carried out on both by Biometric Hash algorithm up to matching score computation. The fusion is carried out by the combination of the matching scores of two instances of one handwritten semantic. The authors demonstrated that when using three semantics and fusion strategies, improvements can be observed in comparison to the best individual results.

Adams Wai-Kin et al. [7] have presented a feature-level coding scheme for improving the performance of PalmCode fusion which applies four Gabor filters to the preprocessed palmprint images to compute four PalmCodes. Karthik Nandakumar et al. [10] have studied the effect of different score normalization techniques in a multimodal biometric system. Min-max, z-score, and tanh normalization techniques followed by a simple sum of scores fusion method result in a higher GAR than all the other normalization and fusion techniques.

## 3   Proposed Method

### 3.1   Choice of Modality

In this paper a hand-based biometric technique, finger-knuckle-print (FKP) have been used, FKP refers to the image pattern of the outer surface around the phalangeal joint of one's finger, which is formed by bending slightly the finger knuckle [4]. The experiments are developed for personal authentication using DZhang FKP database. FKP images were collected from 165 volunteers, including 125 males and 40 females. The database contains FKPs from four types of fingers, left index fingers, left middle fingers, right index fingers and right middle fingers. The DZhang database is available at the website of Biometrics Research Centre, the Hong Kong Polytechnic University.

### 3.2   Pre-Processing

This section describe the Region of Interest (ROI) extraction, the process involved to extract ROI for each instance is as follows. It is necessary and critical to align FKP images by adaptively constructing a local coordinate system for each image. With such a coordinate system, an ROI can be cropped from the original image using the following steps suggested in [4], as shown in Figure-1.

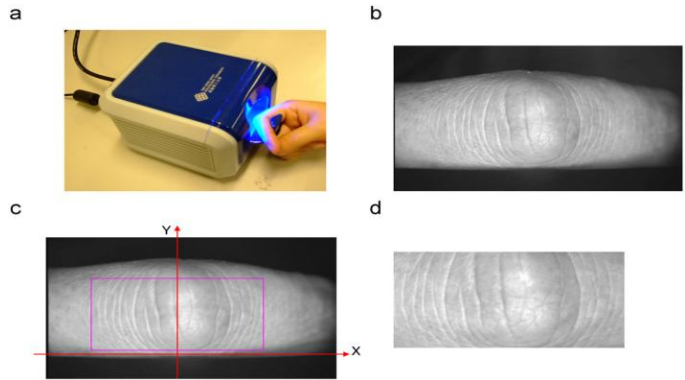

**Fig-1: a)**Image acquisition device is being used to collect FKP samples **b)**Sample FKP image **c)**ROI coordinate system, where the rectangle indicates the area **d)**Extracted ROI

- Image down-sampling.
- Determine the X-axis of the coordinate system.
- Canny edge detection.
- Convex direction coding.
- Determine the Y-axis of the coordinate system.
- Crop the ROI image

### 3.3 Pre-Processing

The Log-Gabor has been used as feature extraction algorithm; it overcomes some of traditional disadvantages found in Gabor filters. Log-Gabor filter was introduced by [5]. Log-Gabor filters basically consist of logarithmic transformation of the Gabor domain which eliminates the annoying DC-component that is allocated in medium and high-pass filters. The response generated by the Log-Gabor is Gaussian; when it is viewed in frequency scale it appears linear. This captures more information of high frequency areas and also exhibits the characteristics of high pass filters (retains information about edges and prominent (sharp) features).

## 5  Result and Discussion

This section deals with the investigation consequences of combining many biometrics at Feature level fusion to measure the performance of multi-instance system. In all the experiments, performance is measured in terms of False Acceptance Rate (FAR in %) and corresponding Genuine Acceptance Rate (GAR in %). First the performance of a single instance biometric system is measured; later the results for multi-instance biometric system are evaluated. The results obtained from single instance biometric system are tabulated in Table-1, and depicted as Receiver Operating Characteristic (ROC) curve in Figure-2.

**Table-1**: The performance of single instances

| FAR % | GAR % | | | |
|---|---|---|---|---|
| | **Right-Index ( RI )** | **Right-Middle ( RM )** | **Left-Index ( LI )** | **Left-Middle ( LM )** |
| 0.01 | 54.66 | 59.11 | 53.34 | 61.56 |
| 0.1 | 66.67 | 70.67 | 64.45 | 70.89 |
| 1 | 77.11 | 80.12 | 78.00 | 81.13 |

From Table-1 it can be observed that the middle finger has a little more score than the index finger for both hands. Next we show the fusion performance results of two and three instances, under the *feature level fusion*.

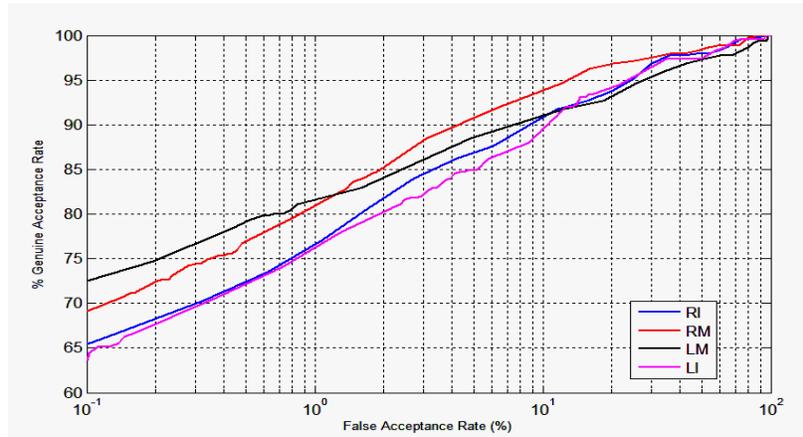

**Fig-2**: The ROC curve performance of single instance.

### 5.1 Biometrics Fusion Strategies

In general a biometric system works in two modes: enrollment and authentication [8]. During the enrollment process a user will be registered within the system by presenting his/her physiological or behavioral biometric trait to acquisition module. Verification and identification are the two modes an authentication can be carried out. While verification an attempt is made to verify the claimed identity of unknown individual "Is this person who he claims to be?", at identification an attempt is made to establish the identity of an individual, "Who is this person?".

Fusion of biometric systems, algorithms and/or traits is a well known solution to improve authentication performance of biometric systems. Researchers have shown that multi-biometrics, i.e., fusion of multiple biometric evidences, enhances the recognition performance [13].

In biometric systems; fusion can be performs at different levels; *Sensor Level, Feature Level, Score Level, and Decision Level Fusion* [11].

### 5.2 Levels of Fusion

- **Sensor Level Fusion** entails the consolidation of evidence presented by multiple sources of raw data before they are subjected to feature extraction. Sensor level fusion can benefit multi-sample systems which capture multiple snapshots of the same biometric.
- **Feature Level Fusion** In feature-level fusion, the feature sets originating from multiple biometric algorithms are consolidated into a single feature set by the application of appropriate feature normalization, transformation, and reduction schemes. The primary benefit of feature-level fusion is the detection of correlated feature values generated by different biometric algorithms and, in the process, identifying a salient set of features that can improve recognition accuracy[3][14]. The block diagram of Score Level Fusion is shown in Figure-3
- **Score Level Fusion**, the match scores output by multiple biometric matchers are combined to generate a new match score (a scalar).
- **Decision Level Fusion**, fusion is carried out at the abstract or decision level when only final decisions are available, this is the only available fusion strategy (e.g. AND, OR, Majority Voting, Weighted Majority Voting, Bayesian Decision Fusion).

In all the experiment, the data have been fused at feature level, using different normalization rules, such as (Min-Max, Median and absolute Median, Z-Score, Tanh) for two and three instances combination of the four fingers.

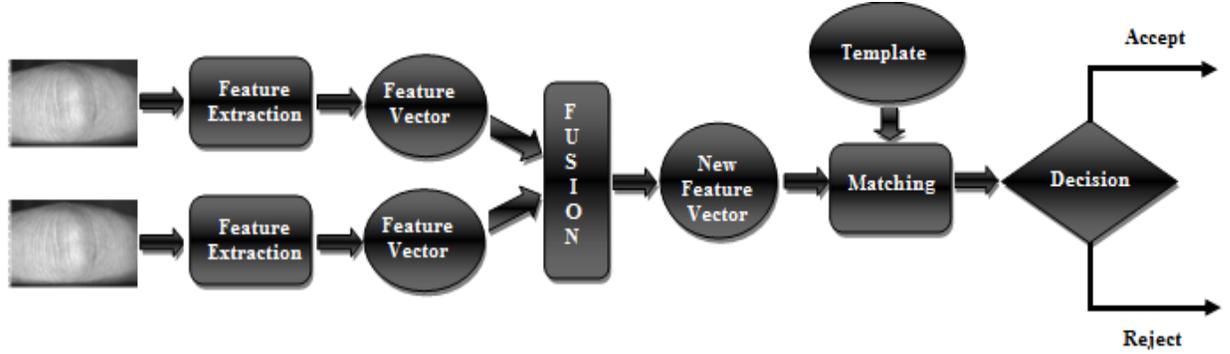

**Fig-3**: The block diagram of Feature Level Fusion

### 5.3 Score Normalization

Score normalization refers to changing the location and scale parameters of the matching score distributions at the output of the individual matchers, so that the scores of different matchers are transformed into a common domain. In a good normalization scheme, the estimates of the location and scale parameters must be robust and efficient. Robustness refers to insensitivity to the presence of outliers. Efficiency refers to the proximity of the obtained estimate to the optimal estimate when the distribution of the data is known [2][3].

- **Min-Max**: The simplest normalization technique is the *Min-Max* normalization. Min-Max normalization is best suited for the case where the bounds (maximum and minimum values) of the scores produced by a matcher are known. In this case, we can easily transform the minimum and maximum scores to 0 and 1 respectively. Let $s_j^i$ denote the $i^{th}$ match score output by the $j^{th}$ matcher, i = 1,2,..., N; j = 1, 2, . . . , R (R is the number of matchers and N is the number of match scores available in the training set). The min-max normalized score $ns_j^i$ for the test score $s_j^i$ is given by[2][3]:

$$ns_j^t = \frac{s_j^t - min_{i=1}^N s_j^i}{max_{i=1}^N s_j^i - min_{i=1}^N s_j^i}$$

- **Z-Score**: The most commonly used score normalization technique is the z-score normalization that uses the arithmetic mean and standard deviation of the training data. This scheme can be expected to perform well if the average and the variance of the score distributions of the matchers are available. If we do not know the values of these two parameters, then we need to estimate them based on the given training set. The z-score normalized score is given by [2][3]:

$$ns_j^t = \frac{s_j^t - \mu_j}{\sigma_j}$$

Where $\mu_j$ is the arithmetic mean and $\sigma_j$ is the standard deviation for the $j^{th}$ matcher. However, both mean and standard deviation are sensitive to outliers and hence, this method is not robust.

- **Median and Median Absolute Deviation (MAD)**: The median and median absolute deviation (MAD) statistics are insensitive to outliers as well as points in the extreme tails of the distribution. Hence, a normalization scheme using median and MAD would be robust and is given by[2][3]:

$$ns_j^t = \frac{s_j^t - med_j}{MAD_j}$$

Where $med_j = median_{i=1}^N s_j^i$ and $MAD = median_{i=1}^N | s_j^i - med_j |$

- **Tanh-Estimator:** The *tanh-estimators* introduced by Hampel et al., 1986 are robust and highly efficient. The tanh normalization is given by[2][3]:

$$ns_j^t = 1/2 \left\{ tanh\left( 0.01 \left( \frac{s_j^t - \mu_j}{\sigma_j} \right) \right) + 1 \right\}$$

This method is not sensitive to outliers. If many of the points that constitute the tail of the distributions are discarded, the estimate is robust but not efficient (optimal). On the other hand, if all the points that constitute the tail of the distributions are considered, the estimate is not robust but its efficiency increases.

Go further the experimental results of the fusion for two and three instances at feature level with different normalization techniques are shown. Table-2 shows the fusion result of two instances and Table-3 shows the fusion result of three instances.

**Table-2:** the performance of two instances at Feature level with different normalization techniques

| FAR (%) | GAR (%) With Z-Score Normalization | | | | | |
|---|---|---|---|---|---|---|
| | RI+RM | RI+LI | RI+LM | LI+LM | RM+LM | RM+LI |
| 0.01 | 68.22 | 56.67 | 65.56 | 58.00 | 70.67 | 63.78 |
| 0.10 | 76.22 | 71.33 | 76.89 | 71.33 | 78.67 | 73.56 |
| 1.00 | 86.22 | 80.89 | 87.33 | 83.78 | 89.33 | 85.33 |
| | GAR (%) With Tanh Normalization | | | | | |
| 0.01 | 68.22 | 58.23 | 65.78 | 64.67 | 70.67 | 64.00 |
| 0.10 | 76.22 | 70.88 | 76.89 | 71.33 | 78.67 | 73.56 |
| 1.00 | 88.22 | 80.89 | 87.33 | 83.78 | 89.33 | 85.33 |
| | GAR (%) With Median Absolute Normalization | | | | | |
| 0.01 | 57.56 | 52 | 58.00 | 55.35 | 52.64 | 51.33 |
| 0.10 | 69.78 | 65.78 | 69.56 | 67.33 | 66.23 | 65.23 |
| 1.00 | 81.33 | 78.22 | 82.00 | 81.56 | 83.34 | 82.67 |
| | GAR (%) With Min-Max Normalization | | | | | |
| 0.01 | 58.00 | 51.87 | 52.22 | 48.23 | 63.78 | 53.33 |
| 0.10 | 69.11 | 62.89 | 66.00 | 65.33 | 74.00 | 67.33 |
| 1.00 | 82.44 | 79.11 | 82.22 | 82.67 | 80.00 | 77.89 |

**Table-3:** the performance of three instances at Feature level with different normalization techniques

| FAR (%) | GAR (%) With Z-Score Normalization | | | |
|---|---|---|---|---|
| | *RI+RM+LI* | *RI+RM+LM* | *RI+LI+LM* | *RM+LI+LM* |
| 0.01 | 54.66 | 59.11 | 53.34 | 61.56 |
| 0.10 | 66.67 | 70.76 | 64.45 | 70.89 |
| 1.00 | 77.11 | 80.12 | 78.00 | 81.13 |
| | GAR (%) With Tanh Normalization | | | |
| 0.01 | 54.66 | 59.11 | 53.34 | 61.56 |
| 0.10 | 66.67 | 70.76 | 64.45 | 70.89 |
| 1.00 | 77.11 | 80.12 | 78.00 | 81.13 |
| | GAR (%) With Median and MAD Normalization | | | |
| 0.01 | 54.66 | 59.11 | 53.34 | 61.56 |
| 0.10 | 66.67 | 70.76 | 64.45 | 70.89 |
| 1.00 | 77.11 | 80.12 | 78.00 | 81.13 |

|        | GAR (%) With Min-Max Normalization |       |       |       |
|--------|-------|-------|-------|-------|
| 0.01   | 54.66 | 59.11 | 53.34 | 61.56 |
| 0.10   | 66.67 | 70.76 | 64.45 | 70.89 |
| 1.00   | 77.11 | 80.12 | 78.00 | 81.13 |

From Table-2 it can be observed that the fusion of two instances finger has a significant improve score over the single instance with *Z-score* and *tanh-estimators*, but does not have much improvement with *Min-Max* and *Median & MAD*. Figure-4 shows the ROC curve for the performance fusion of two instances at Feature level with different normalization techniques. From Table-3 it can be observed that the fusion of three instances does not have any score improvement over the fusion of two instances, even worst performance depending upon the combinations of the instances. Figure-5 shows the ROC curve for the performance of the fusion of three instances at Feature level with different normalization techniques.

# 6 Conclusion

Analyzing the performance of using FKP images as a biometric has been done. Four matching score normalization techniques experimentally evaluated to improve the performance fusions of different instances. The results of evaluation represented by means of system performance (expressed by ROC, FAR vs GAR). From the analysis of experimental results and observations it can be concluded that a multi-instance biometric fusion is given better performance than single instances. This shows that using multiple instance of biometric which collected using single sensor; can have the security level. However, from the experimental results and observations the degree of improvement in accuracy by fusing multiple instances is marginal. Since different instance of the same instance produces the redundant features.

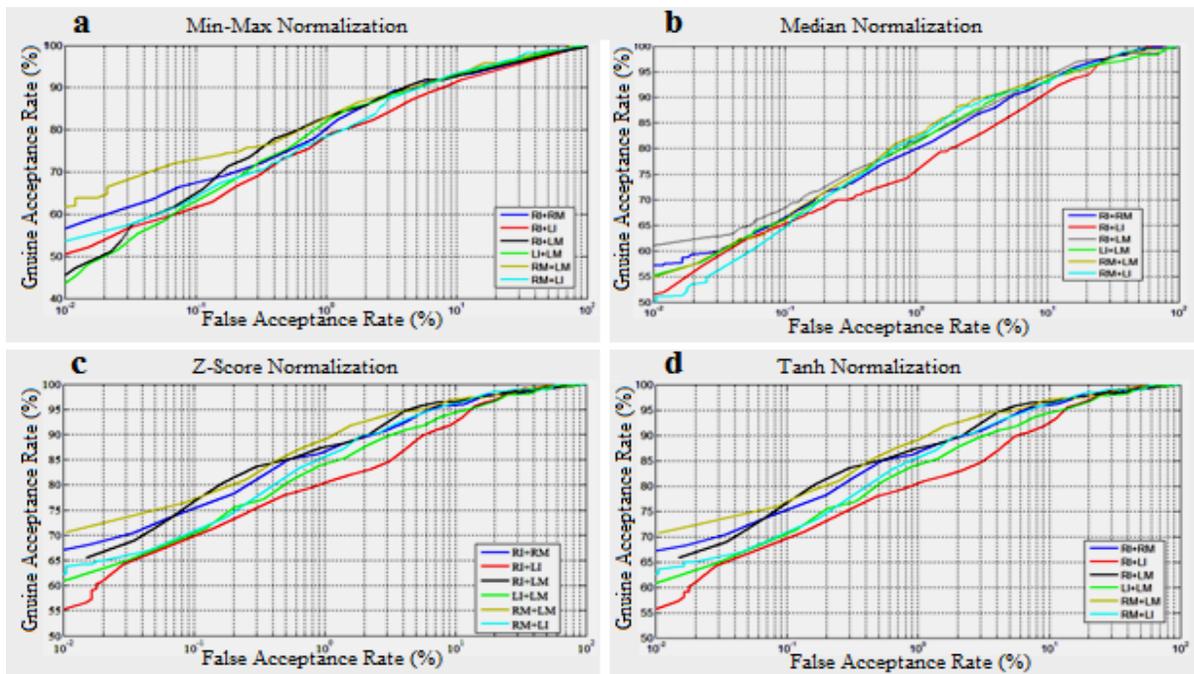

**Fig-4**: The ROC curve Feature Level Fusion combination of two instances with different Normalization rules, **a)** Min-Max  **b)** Median  **c)** Z-Score  **d)** Tanh.

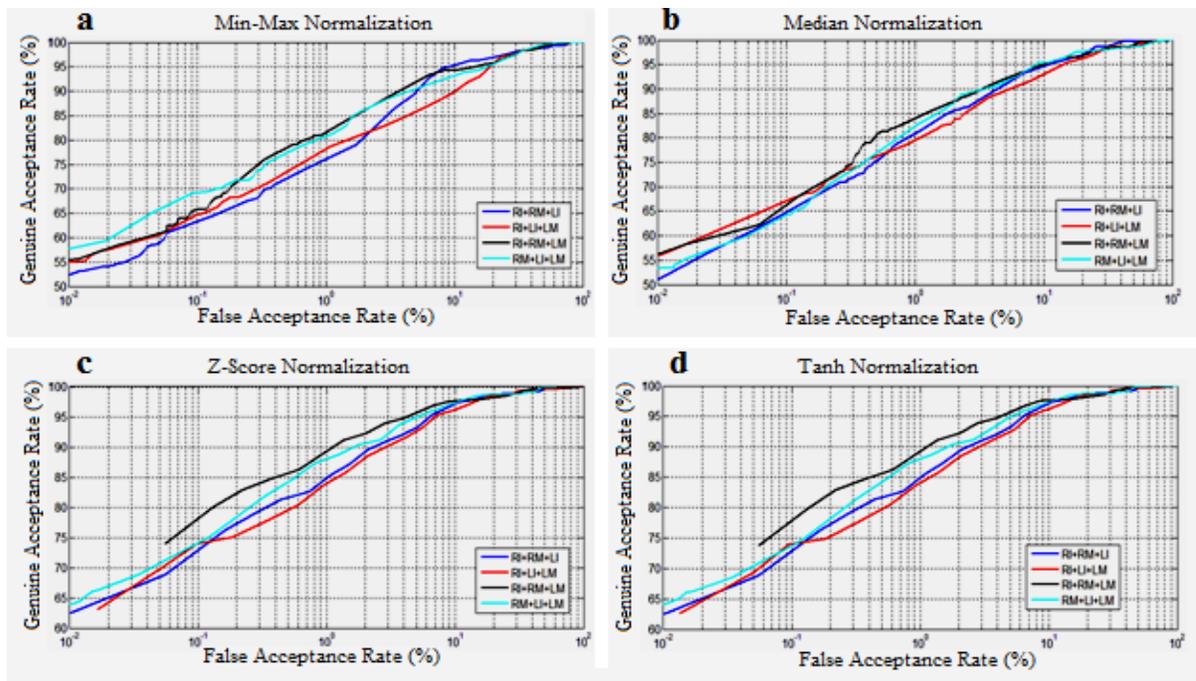

**Fig-5**: The ROC curve Feature Level Fusion combination of three instances with different Normalization rules, **a)** Min-Max  **b)** Median  **c)** Z-Score  **d)** Tanh.